\pdfoutput=1

\documentclass[11pt]{article}

\usepackage[]{acl}

\usepackage{multirow}
\usepackage{times}
\usepackage{latexsym}
\usepackage{soul}
\usepackage{url}
\usepackage{bm}
\usepackage{hyperref}
\usepackage[utf8]{inputenc}
\usepackage{graphicx}
\usepackage{amsmath}
\usepackage{mathrsfs}
\usepackage{amsthm}
\usepackage{booktabs}
\usepackage{algorithm}
\usepackage{algorithmic}
\usepackage{booktabs}
\usepackage{amssymb}
\usepackage{color}
\usepackage{multicol}
\usepackage{subfigure}
\usepackage{array}
\usepackage{color}
\graphicspath{{figure/}}
\usepackage[T1]{fontenc}

\usepackage[utf8]{inputenc}

\usepackage{microtype}

\title{Zero-Shot Chain-of-Thought Reasoning Guided by Evolutionary Algorithms in Large Language Models}

\author{Feihu Jin$^{1,2}$, Yifan Liu$^{1,2}$ \and Ying Tan$^{1,2,3,4}$ \thanks{~ Corresponding Author}\\
  $^1$ School of Intelligence Science and Technology, Peking University\\
  $^2$ Institute for Artificial Intelligence, Peking University \\
  $^3$ National Key Laboratory of General Artificial Intelligence\\ 
  $^4$ Key Laboratory of Machine Perceptron (MOE), Peking University\\
\text{fhjin@stu.pku.edu.cn}, \text{liuyifan731@163.com}, \text{ytan@pku.edu.cn}}

\begin{document}
\maketitle
\begin{abstract}
Large Language Models (LLMs) have demonstrated remarkable performance across diverse tasks and exhibited impressive reasoning abilities by applying zero-shot Chain-of-Thought (CoT) prompting. However, due to the evolving nature of sentence prefixes during the pre-training phase, existing zero-shot CoT prompting methods that employ identical CoT prompting across all task instances may not be optimal. In this paper, we introduce a novel zero-shot prompting method that leverages evolutionary algorithms to generate diverse promptings for LLMs dynamically. Our approach involves initializing two CoT promptings, performing evolutionary operations based on LLMs to create a varied set, and utilizing the LLMs to select a suitable CoT prompting for a given problem. Additionally, a rewriting operation, guided by the selected CoT prompting, enhances the understanding of the LLMs about the problem. Extensive experiments conducted across ten reasoning datasets demonstrate the superior performance of our proposed method compared to current zero-shot CoT prompting methods on GPT-3.5-turbo and GPT-4. Moreover, in-depth analytical experiments underscore the adaptability and effectiveness of our method in various reasoning tasks\footnote{We have released our code in: \url{https://github.com/stan-anony/Zero-shot-EoT-Prompting}}.
\end{abstract}

\section{Introduction}
The capacity for logical inference stands out as a defining characteristic of human intelligence, granting us the ability to engage in deduction, induction, and problem-solving. With the revolutionary advancement of pre-training \cite{Brown2020LanguageMA, DBLP:journals/corr/abs-2302-13971, chatgpt, Achiam2023GPT4TR}, the rise of large language models (LLMs) has firmly established itself as a cornerstone in the field of natural language processing (NLP), showcasing exceptional performance across a spectrum of NLP tasks. However, LLMs often face challenges in the nuanced domain of reasoning, prompting researchers to strategically leverage their embedded knowledge through the conditioning of LLMs on a limited set of illustrative examples, referred to as few-shot learning \cite{NEURIPS2022_9d560961,wang2023selfconsistency}, or through the provision of prompts for solving problems in the absence of illustrative examples, constituting a paradigm known as zero-shot learning \cite{NEURIPS2022_8bb0d291}.

Current research is predominantly focused on designing diverse prompting strategies to guide the reasoning processes of LLMs. For instance, \citet{NEURIPS2022_9d560961} propose the few-shot Chain-of-Thought prompting, involving the use of a limited number of manually demonstrated reasoning examples to enable LLMs to explicitly generate intermediate reasoning steps before predicting the final answer. Various approaches have been explored to eliminate the need for manually selected examples in few-shot CoT prompting. For instance, \citet{NEURIPS2022_8bb0d291} introduce zero-shot CoT prompting by appending \emph{"Let's think step by step"} to the target problem, PS+ prompting \citet{wang-etal-2023-plan} add \emph{"Let's first understand the problem and devise a plan to solve the problem. Then, let's carry out the plan and solve the problem step by step"} after the target problem, and RE2 prompting \citet{Xu2023ReReadingIR} add \emph{"Read the question again"} combined with \emph{"Let's think step by step"} to the target problem. However, these zero-shot CoT prompting methods employ uniform CoT prompting across all instances of the task. Given the ongoing evolution of sentence prefixes during the pre-training phase of extensive language models, using identical CoT prompting for all instances may introduce disruptions to predictive accuracy and potentially result in a degradation of overall performance. Consequently, a fundamental query emerges: Is it feasible to ascertain an appropriate CoT prompting for each instance within a discrete space?

Fortunately, evolutionary algorithms (EA) \cite{DBLP:books/daglib/0019083,Hansen2003ReducingTT,8239679} provide a solution. EA represents a category of optimization algorithms inspired by the principles of natural evolution. The steps involving crossover, mutation, and selection in EA can generate various CoT promptings. In this paper, we introduce a novel zero-shot CoT prompting method guided by evolutionary algorithms named zero-shot EoT prompting. The process begins by initializing two CoT promptings. Using Large Language Models as the optimizer within an evolutionary algorithm framework, we perform crossover and mutation operations on the initialized CoT promptings, generating a diverse set of new CoT promptings. Subsequently, the LLMs are utilized to select a CoT prompting that is deemed suitable for the current problem. Furthermore, to deepen the understanding of LLMs of the current problem, a rewriting operation is performed on the selected CoT prompting. The LLMs engage in reasoning based on the rewritten problem. This strategy aims to capitalize on the diversity of CoT prompting generated through the EA, combined with problem rewriting, to provide richer information that encourages the LLMs to attain a more profound understanding of the given problem.

To validate the effectiveness of our proposed zero-shot EoT prompting, we conduct a comprehensive series of experiments across ten datasets, covering arithmetic, commonsense, and symbolic reasoning. The experiments are carried out on GPT-3.5-turbo and GPT-4. The results indicate that our zero-shot EoT prompting outperforms existing zero-shot CoT prompting and PS+ prompting methods across all reasoning datasets. Its comparable performance to few-shot CoT prompting is particularly noteworthy, especially in arithmetic and symbolic reasoning. Additionally, extensive analytical experiments are conducted to gain a deeper understanding of the different components of zero-shot EoT prompting and the impact of various factors on EoT prompting.

\section{Preliminaries}
\subsection{Zero-shot Chain-of-Thought Prompting} 
In-context learning leverages a few demonstrations as a prompt and conducts inference without training the model parameters \cite{Brown2020LanguageMA}. Chain-of-thought (CoT) prompting \cite{NEURIPS2022_9d560961} has been proposed as a type of in-context learning that decomposes the original problem into several small parts and achieves encouraging results on many complex reasoning tasks in large language models. Moreover, the zero-shot chain-of thoughts prompting \cite{NEURIPS2022_8bb0d291} has shown impressive effectiveness on various tasks in large language models by attaching a sentence before the reasoning process. For standard zero-shot CoT prompting, given the reasoning question $\mathcal{Q}$, zero-shot CoT specific instructions $\mathcal{T}$ like \emph{"Let's think step by step."}, we formalize this simple yet fundamental solving paradigm as:
\begin{equation}
\begin{aligned}
P(\mathcal{A}|\mathcal{T},\mathcal{Q}) = P(\mathcal{A}|\mathcal{T},\mathcal{Q},\mathcal{C})P(\mathcal{C}|\mathcal{T},\mathcal{Q})
\end{aligned}
\end{equation}
where $\mathcal{C}$ denotes a sampled rationale in natural language and $\mathcal{A}$ is the generated answer. As such, LLMs can perform complex reasoning by decomposing the problem into sequential or modular steps.
\begin{figure*}[bt]
    \centering
    \includegraphics[scale=0.43]{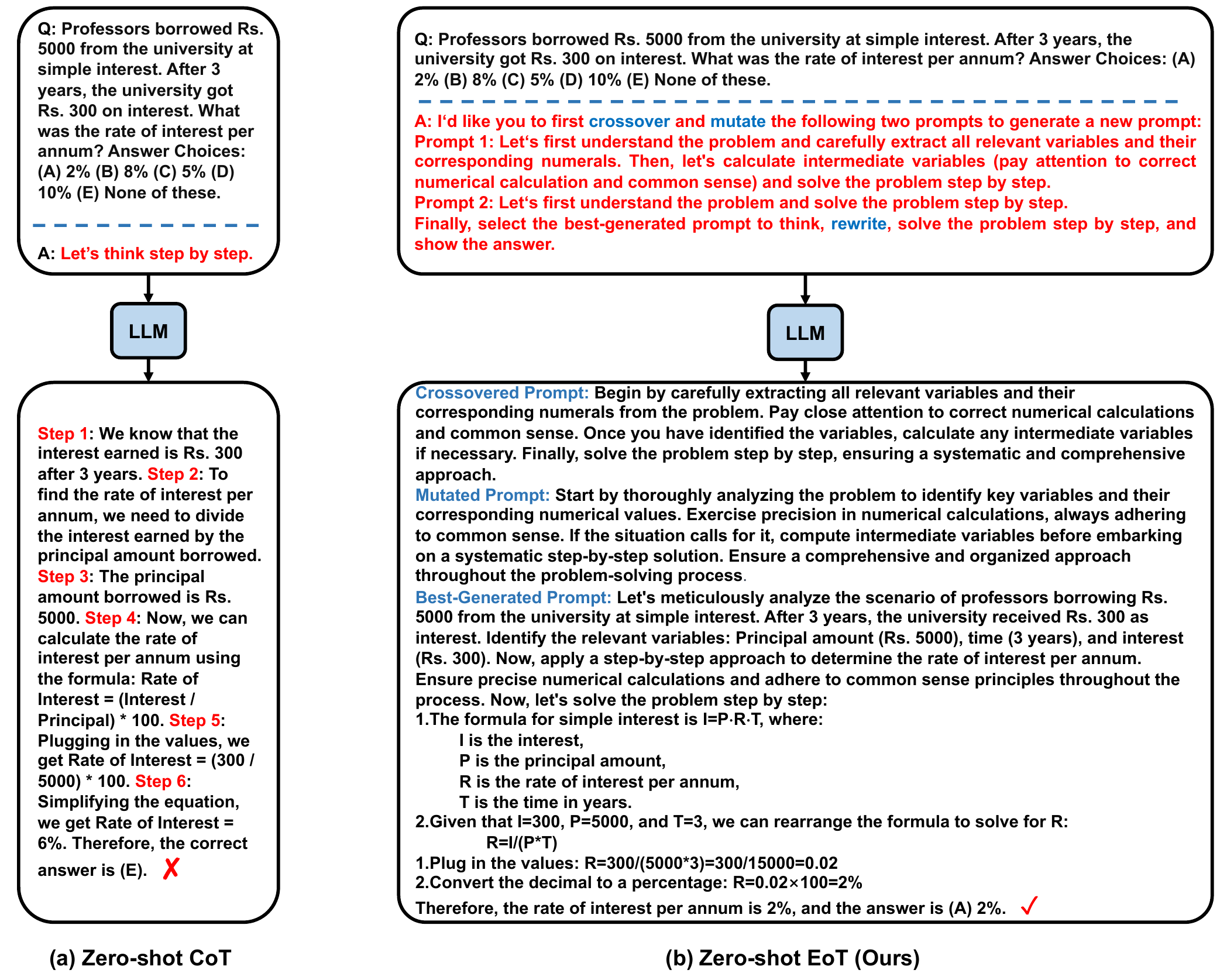}
    \caption{Example inputs and outputs of GPT-3.5-turbo with (a) Zero-shot CoT prompting and (b) Zero-shot EoT prompting. Zero-shot CoT prompting attaches the sentence \emph{"Let's think step by step"} for each instance to encourage LLMs to generate multi-step reasoning. Our proposed method, EoT prompting, uses the LLMs as an evolutionary optimizer and generates suitable CoT prompting for each instance.}
    \label{model-arch}
\end{figure*}
\section{Method}
\noindent\textbf{Overview.} We introduce our proposed zero-shot EoT prompting. EoT utilizes Large Language Model as an evolutionary optimizer, performing crossover and mutation operations on two given promptings to generate diverse promptings. Subsequently, EoT empowers LLMs to autonomously select the most suitable or optimal prompting from the generated set as the final prompting. Finally, EoT employs the chosen prompt to instruct LLMs in rewriting given problems, generating an intermediate reasoning process, and predicting the final answer for the input problem. Similar to zero-shot CoT prompting, our EoT prompting incorporates an answer extraction prompting, such as "Therefore, the answer (in Arabic numerals) is," to extract the answer for evaluation.
\subsection{Prompt Generation Through Evolutionary Algorithms}
As depicted in Figure \ref{model-arch}(a), zero-shot CoT prompting \cite{NEURIPS2022_8bb0d291} appends the same sentence \emph{"Let's think step by step"} or the recently proposed Plan-and-Solve prompting \cite{wang-etal-2023-plan} connects the same sentence \emph{"Let's first understand the problem and devise a plan to solve the problem. Then, let's carry out the plan and solve the problem step by step"} to each instance, encouraging LLMs to generate multi-step reasoning. Given the continuous evolution of sentence prefixes during the pre-training phase of large language models, using identical CoT prompting for all instances may disrupt predictions and lead to a decline in performance.

To address these concerns, we aim to identify suitable CoT prompting for each instance of the current reasoning task within a discrete space before proceeding with the reasoning process. However, determining the most suitable CoT prompting for each instance in a discrete space poses a challenge. Fortunately, evolutionary algorithms provide a solution. We employ the large language model as an evolutionary optimizer, executing crossover and mutation on the initialized CoT prompting, denoted as LLM-Crossover and LLM-Mutation. As illustrated in Figure \ref{model-arch}(b), for a given problem $\mathcal{Q}$, we initialize two CoT promptings $\mathcal{T}_1$ and $\mathcal{T}_2$. Subsequently, we first use the large language model as the evolutionary optimizer, applying the LLM-Crossover operation on $\mathcal{T}_1$ and $\mathcal{T}_2$, which is defined as:
\begin{equation}
\begin{aligned}
\mathcal{T}_c = \text{LLM-Crossover}(\mathcal{T}_1, \mathcal{T}_2)
\end{aligned}
\end{equation}
Then, we enable LLM-Mutation on the crossovered CoT prompting $\mathcal{T}_c$, which is defined as:
\begin{equation}
\begin{aligned}
\mathcal{T}_m = \text{LLM-Mutation}(\mathcal{T}_c)
\end{aligned}
\end{equation}
This leverages the powerful generative capability of the large language model to generate additional CoT promptings. 

Pursuing a more diverse set of selectable CoT promptings, it is customary to subject the model to crossover and mutation operations iteratively. However, the temporal demand tends to escalate proportionally with the quantity of generated CoT promptings. Consequently, we opt for a default strategy of conducting a singular round of crossover and mutation operations to mitigate reasoning time. As illustrated in Figure \ref{figure_popsize}, our analysis delves into the correlation between the number of CoT promptings (i.e., the population size $N$) generated through multiple rounds of crossover and mutation operations and the performance of LLMs.

\subsection{Problem Rewriting with Generated Prompt and Answer Extraction}
Based on the generated and initialized pool of CoT promptings, we enable the LLMs to select the most optimal or contextually suitable CoT prompting for the current problem $\mathcal{Q}$. Subsequently, to enhance the retention of the LLMs regarding the problem, we employ the selected CoT prompting to rewrite the question $\mathcal{Q}$ and instruct the LLMs to conduct reasoning. The formalization of this process is exemplified as follows:
\begin{equation}
\begin{aligned}
P(\mathcal{A}|\mathcal{T}_o,\mathcal{Q}) = P(\mathcal{A}|\mathcal{T}_o,R(\mathcal{Q}),\mathcal{C})P(\mathcal{C}|\mathcal{T}_o,R(\mathcal{Q}))
\end{aligned}
\end{equation}
Here, $\mathcal{T}_o$ denotes the selected CoT prompting by LLMs, $\mathcal{C}$ denotes a sampled rationale in natural language, $\mathcal{A}$ is the generated answer, and $R(\cdot)$ means rewriting the question $\mathcal{Q}$ with $\mathcal{T}_o$.
For instance, in Figure \ref{model-arch}b, for a given question $\mathcal{Q}$: \emph{Professors borrowed Rs. 5000 from the university at simple interest. After 3 years, the university got Rs. 300 on interest. What was the rate of interest per annum? Answer Choices: (A) 2\% (B) 8\% (C) 5\% (D) 10\% (E) None of these.} We employ the chosen CoT prompting to rewrite the question $R(\mathcal{Q})$: \emph{Let's meticulously analyze the scenario of professors borrowing Rs. 5000 from the university at simple interest. After 3 years, the university received Rs. 300 as interest. Identify the relevant variables: Principal amount (Rs. 5000), time (3 years), and interest (Rs. 300). Now, apply a step-by-step approach to determine the rate of interest per annum. Ensure precise numerical calculations and adhere to common sense principles throughout the process.} Then, the LLMs generate an intermediate reasoning process and predict the final answer for the question $\mathcal{Q}$. Moreover, our method defaults to employing the greedy decoding strategy for the generation of output. 

Similar to the zero-shot CoT prompting, our EoT prompting incorporates specific trigger sentences, such as \emph{"Therefore, the answer (arabic numerals) is"}, into the sentences generated by LLMs through EoT prompting. Following this augmentation, the composite text is reintroduced to LLMs, producing the desired answer format. In Appendix \ref{answer_ext}, we present the trigger sentences utilized for different reasoning tasks.

\section{Experiments}
\subsection{Experimental Setup}
\noindent \textbf{Datasets} \
We systematically evaluate the efficacy of our proposed method across ten datasets encompassing three main categories: arithmetic, commonsense, and symbolic tasks. For arithmetic reasoning tasks, we consider the following six arithmetic reasoning problem benchmarks: (1) MultiArith \cite{roy-roth-2015-solving}, (2) GSM8K \cite{Cobbe2021TrainingVT}, (3) AddSub \cite{hosseini-etal-2014-learning}, (4) AQuA \cite{ling-etal-2017-program}, (5) SingleEq \cite{koncel-kedziorski-etal-2015-parsing}, and (6) SVAMP \cite{patel-etal-2021-nlp}. SingleEq and AddSub comprise more straightforward problems that do not require multi-step task resolution calculations. Conversely, MultiArith, AQUA, GSM8K, and SVAMP present more intricate challenges, demanding multi-step reasoning for effective problem-solving. In the realm of commonsense reasoning, we include (7) CommonsenseQA \cite{talmor-etal-2019-commonsenseqa} and (8) StrategyQA \cite{geva-etal-2021-aristotle}. CommonsenseQA requires the application of diverse forms of commonsense knowledge for accurate answers. Meanwhile, StrategyQA tasks models with deducing implicit multi-hop reasoning to respond to posed questions. For symbolic tasks, we select Last Letter Concatenation and Coin Flip \cite{NEURIPS2022_9d560961}. Last Letter Concatenation challenges the model to concatenate the last letters of individual words. At the same time, the Coin Flip task requires the model to determine whether a coin remains in a heads-up position after being flipped or left undisturbed. Details on dataset statistics are provided in Appendix \ref{dataset}.

\noindent \textbf{Baselines} We conduct a comparative analysis between our proposed zero-shot EoT prompting method and several leading zero-shot CoT prompting methods: (1) Zero-shot CoT prompting \cite{NEURIPS2022_8bb0d291}, which appends a sentence \emph{"Let's think step by step"} before the reasoning process; (2) Zero-shot PS and PS+ prompting \cite{wang-etal-2023-plan}, employing a "plan-and-solve" strategy to guide the model throughout the inference process; (3) Zero-shot RE2 prompting \cite{Xu2023ReReadingIR}, a plug \& play approach that entails re-reading the question before engaging in the reasoning process. Additionally, we compare our method with two few-shot CoT prompting methods: Few-shot Manual-CoT prompting \cite{NEURIPS2022_9d560961}, utilizing eight manually crafted examples as demonstrations, and Few-shot AuTo-CoT prompting \cite{zhang2023automatic}, which automatically selects examples through clustering for diversity.

\noindent \textbf{Implementation Details}
We mainly use ChatGPT (GPT-3.5-turbo-0613) \cite{chatgpt} as the backbone language model. Regarding decoding strategy, we employ greedy decoding with a temperature setting of 0 and implement self-consistency prompting with a temperature setting of 0.7. Furthermore, to fortify the robustness and generalizability of our proposed method, we conduct complementary evaluations utilizing GPT-4 \cite{Achiam2023GPT4TR}. For the few-shot baselines, Manual-CoT and Auto-CoT, we adhere to the configurations outlined in the \citet{NEURIPS2022_9d560961} and \cite{zhang2023automatic}. We adopt accuracy as our evaluation metric for all datasets.

\begin{table*}[!htbp]
\centering
\scriptsize
\caption{Accuracy of six math reasoning datasets on GPT-3.5-turbo with different zero-shot and few-shot CoT prompting methods. The boldfaced and underlined fonts indicate the best and the second results in the zero-shot settings, respectively.}
\resizebox{\textwidth}{!}{%
\renewcommand{\arraystretch}{1.3}
\begin{tabular}{@{}lccccccc@{}}
\toprule
Method & 
  {\begin{tabular}[c]{@{}c@{}}MultiArith\end{tabular}} &
  {\begin{tabular}[c]{@{}c@{}}GSM8K\end{tabular}} &
  {\begin{tabular}[c]{@{}c@{}}AddSub\end{tabular}} &
  {\begin{tabular}[c]{@{}c@{}}AQuA\end{tabular}} &
  {\begin{tabular}[c]{@{}c@{}}SingleEq\end{tabular}} &
  {\begin{tabular}[c]{@{}c@{}}SVAMP\end{tabular}} &
  {\begin{tabular}[c]{@{}c@{}}Average \end{tabular}}
  \\ \midrule
Zero-shot CoT    & 95.3  &75.3  & 86.5        & 55.3        & \underline{92.9} & 79.0    &80.7 \\
Zero-shot PS   & 92.4 & 76.3 & 85.7        & 56.7         & 90.1 & 75.8       & 79.5  \\
Zero-shot PS+  & 93.8   & 76.1& 86.6        & 58.9         & 92.5& 79.4  & 81.2   \\ 
Zero-shot RE2   & \textbf{96.9}    & \textbf{76.9} & \underline{88.7}   & \underline{59.9}     &91.8        & \underline{79.7} & \underline{82.3}     \\ 
Zero-shot EoT (ours)  & \underline{96.4}    & \underline{76.8}  & \textbf{91.1}  & \textbf{62.2}  &\textbf{93.5} & \textbf{81.2}    &\textbf{83.5} \\  \midrule
Few-shot Manual-CoT   & 95.4 & 75.9 & 89.9        & 58.7         & 92.3 & 81.1       & 82.2  \\
Few-shot AuTo-CoT   & 96.2 & 77.3 & 90.7        & 61.7         & 92.7 & 81.8       & 83.4 \\ 
\bottomrule
\end{tabular}
}
\label{table_math}
\end{table*}

\subsection{Main Results}
\noindent\textbf{Results on Arithmetic Reasoning.}
Table \ref{table_math} presents a thorough performance comparison between our proposed method EoT and existing zero-shot and few-shot baselines on the arithmetic reasoning datasets with GPT-3.5-turbo. In contrast to prevalent zero-shot CoT, PS, and PS+ prompting methodologies, our EoT prompting exhibits notable improvements in performance across six arithmetic reasoning datasets, showcasing particularly significant improvements on the AddSub, SVAMP, AQuA, and SingleEq datasets. Furthermore, on average, our EoT prompting achieves a 2.8\% and 2.3\% score improvement over zero-shot CoT prompting and PS+ prompting methods, respectively. Concerning the zero-shot RE2 prompting, our EoT prompting outperforms it across four datasets, displaying comparable performance on the MultiArith and GSM8K datasets with marginal differentials. The observed similarity between the zero-shot RE2 prompting, characterized by repetitive questions, and our approach of rewriting questions using CoT prompting generated via evolutionary algorithms suggests the advantageous impact of enhancing the model's capacity to retain questions on the reasoning process. Concurrently, we compare our proposed EoT prompting with a few-shot methods: Manual-CoT and Auto-CoT. The results indicate that our proposed EoT prompting surpasses Manual-CoT and Auto-CoT on six and four arithmetic reasoning datasets, respectively, suggesting the effectiveness of our zero-shot EoT prompting in achieving comparable results to few-shot methods in arithmetic reasoning datasets without the need for example selection.

\begin{table}[!htbp]
\centering
\small
\caption{Accuracy of two commonsense reasoning datasets on GPT-3.5-turbo with different zero-shot and few-shot CoT prompting methods. CSQA denotes CommonsenseQA}
\renewcommand{\arraystretch}{1.3}
\begin{tabular}{@{}lcc@{}}
\toprule
Method & CSQA & StrategyQA
  \\ \midrule
Few-shot Manual-CoT   & 75.3 & 70.1  \\
Few-shot AuTo-CoT   & 77.1 & 71.3 \\ \midrule
Zero-shot CoT    & 64.9  &65.7  \\
Zero-shot PS   &68.6 & 66.4 \\
Zero-shot PS+  & 70.9   & 67.8 \\
Zero-shot RE2   & 71.5    & 68.1     \\ 
Zero-shot EoT (ours)  & \textbf{72.7}   & \textbf{69.9}   \\ \bottomrule
\end{tabular}
\label{table_commonsense}
\end{table}
\noindent\textbf{Results on Commonsense Reasoning.}
Table \ref{table_commonsense} shows the result on two commonsense reasoning datasets. In the zero-shot setting, our EoT prompting exhibits superior performance relative to zero-shot CoT prompting, PS prompting, PS+ prompting, and the RE2 prompting methods on two commonsense reasoning datasets. Conversely, compared to two few-shot methods, Manual-CoT and Auto-CoT, our zero-shot EoT prompting demonstrates comparatively lower performance on these two commonsense reasoning datasets. This observation implies that commonsense reasoning problems may necessitate a certain degree of demonstrations to guide the model reasoning process.

\begin{table}[!htbp]
\centering
\small
\caption{Accuracy of two symbolic reasoning datasets on GPT-3.5-turbo with different zero-shot and few-shot CoT prompting methods.}
\renewcommand{\arraystretch}{1.3}
\begin{tabular}{@{}lcc@{}}
\toprule
Method & Last Letters & Coin Flip
  \\ \midrule
Few-shot Manual-CoT   & 75.7 &  99.3 \\
Few-shot AuTo-CoT   & 76.3 & 99.7 \\ \midrule
Zero-shot CoT    & 72.6  &98.6 \\
Zero-shot PS   &71.3 & 96.9 \\
Zero-shot PS+  & 70.4   & 97.6 \\
Zero-shot RE2   & 74.3    & 97.7    \\ 
Zero-shot EoT (ours)  &\textbf{76.8}    & \textbf{98.9}   \\ \bottomrule
\end{tabular}
\label{table_symbolic}
\end{table}

\noindent\textbf{Results on Symbolic Reasoning.}
Table \ref{table_symbolic} shows the result on two symbolic reasoning datasets: Last Letters and Coin Flip. Our EoT prompting demonstrates superior performance compared to zero-shot CoT prompting, PS prompting, PS+ prompting, and the RE2 prompting methods on these two symbolic reasoning datasets, especially in the Last Letter dataset. In contrast to few-shot methods, Manual-CoT, and Auto-CoT, our EoT prompting excels relative to these methods in the Last Letter dataset while demonstrating comparable performance in the Coin Flip dataset. This observation suggests the effectiveness of our zero-shot EoT prompting in achieving comparable results to few-shot methods in symbolic reasoning datasets without the need for example selection. 

\begin{table}[!htbp]
\centering
\small
\caption{Results of different methods measured on three math reasoning datasets with GPT-4.}
\renewcommand{\arraystretch}{1.3}
\begin{tabular}{@{}lccc@{}}
\toprule
Method & AQuA & AddSub & SVAMP
  \\ \midrule
Zero-shot CoT    & 72.8  & 95.2 & 88.7 \\
Zero-shot PS+  & 73.2   & 96.4 & 89.2 \\
Zero-shot RE2   & 73.9    & 96.2 & 90.1    \\ \midrule
Zero-shot EoT (ours)  &\textbf{75.9}    & \textbf{97.7}   & \textbf{92.7}\\ \bottomrule
\end{tabular}
\label{table_gpt4}
\end{table}
\section{Additional Experiments and Analysis}
\subsection{Results of EoT Prompting in GPT-4}
To evaluate the performance of our proposed zero-shot EoT prompting with more powerful models, as shown in Table \ref{table_gpt4}, we conduct experiments on GPT-4 using three arithmetic reasoning datasets: AQuA, AddSub, and SVAMP. We compare our zero-shot EoT prompting against three alternative methods: zero-shot CoT prompting, PS+ prompting, and RE2 prompting. The results presented in Table \ref{table_gpt4} reveal that our zero-shot EoT prompting yields superior performance compared to the three methods, suggesting that our proposed method maintains robust performance advantages when applied to more powerful language models.

\subsection{Ablation Study of EoT}
We perform the ablation study of our EOT prompting measured on four math reasoning datasets under the zero-shot setting to understand the importance of different factors. As delineated in Table \ref{table_ablation}, the notations 'R', 'C', and 'M' denote the operations of rewrite, crossover, and mutate, respectively. Our observations indicate that refraining from employing EoT prompting for problem rewriting results in a discernible decline in model performance across all tasks. This underscores the importance of augmenting the model's comprehension of problems through a more profound engagement, thereby fostering more effective inference. Furthermore, in the course of generating our EoT prompts, the omission of crossover or mutation processes results in a significant performance decrease across all tasks except the SVAMP dataset. Notably, the AQuA dataset exhibits a pronounced performance degradation, emphasizing the indispensability of the crossover and mutation processes in the effective generation of our EoT prompting.
\begin{table}[!htbp]
\centering
\small
\caption{Ablation study of EoT measured on four math reasoning datasets with GPT-3.5-turbo. 'R', 'C', and 'M' denote rewrite, crossover, and mutate, respectively.}
\renewcommand{\arraystretch}{1.3}
\begin{tabular}{@{}lcccc@{}}
\toprule
Method & AQuA & AddSub & SVAMP &GSM8K
  \\ \midrule
EoT    & 62.2 & 91.1 & 81.2  & 76.8\\ 
\quad\ -w/o R  & 61.7   & 90.3 & 80.9 &  75.7\\
\quad\ -w/o C  & 57.8    & 88.9 & 82.5   &  75.8\\ 
\quad\ -w/o M  &55.9    & 87.3  & 81.4 & 76.3\\ \bottomrule
\end{tabular}
\label{table_ablation}
\end{table}

\subsection{Results of Prompting with Self-Consistency}
Existing research suggests that the CoT prompting method can be enhanced through the incorporation of self-consistency \cite{wang2023selfconsistency}. This is achieved through the generation of N reasoning results, with the final answer determined by means of a majority voting process. Our interest is additionally piqued by the prospect of further augmenting the proposed EoT prompting through self-consistency. Consequently, experimental validations are conducted across four arithmetic reasoning datasets: AddSub, AQuA, SingleEq, and SVAMP. As depicted in Table \ref{table_self-consistency}, the comparative assessment involves an analysis of the performance of zero-shot CoT prompting, PS+ prompting, and RE2 prompting subsequent to the application of the self-consistency method. It is discernible that our EoT prompting exhibits superior performance across diverse arithmetic reasoning datasets when compared to these baselines.
\begin{table}[!htbp]
\centering
\small
\caption{Results of different methods in a zero-shot setting with self-consistency measured on four math reasoning datasets with GPT-3.5-turbo.}
\renewcommand{\arraystretch}{1.3}
\begin{tabular}{@{}lcccc@{}}
\toprule
Method & AddSub & AQuA & SingleEq & SVAMP
  \\ \midrule
CoT +SC   & 87.1  & 62.5 & 94.6 & 80.6 \\
PS+ +SC  & 88.5   & 63.1  & 93.7 & 81.1 \\
RE2 +SC  & 89.7    & 63.5 & 94.9 &80.8    \\ \midrule
EoT +SC (ours)  &\textbf{91.9}    & \textbf{65.8}   & \textbf{95.2} &\textbf{82.7} \\ \bottomrule
\end{tabular}
\label{table_self-consistency}
\end{table}

\subsection{Effect of Population Size}
\begin{figure}[!htbp]
\centering
\subfigure{
\begin{minipage}[t]{0.45\linewidth}
\centering
\includegraphics[scale=0.27]{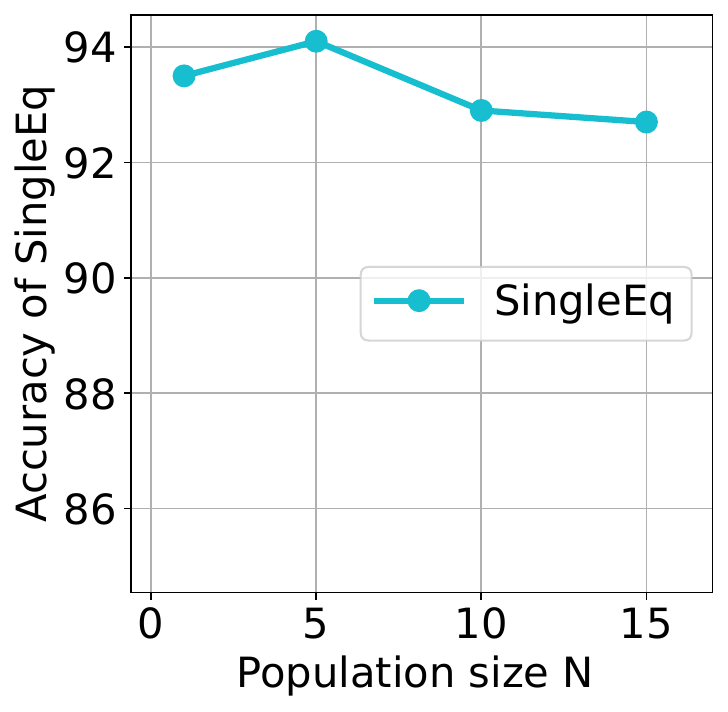}
\label{singleeq}
\end{minipage}%
}%
\subfigure{
\begin{minipage}[t]{0.45\linewidth}
\centering
\includegraphics[scale=0.27]{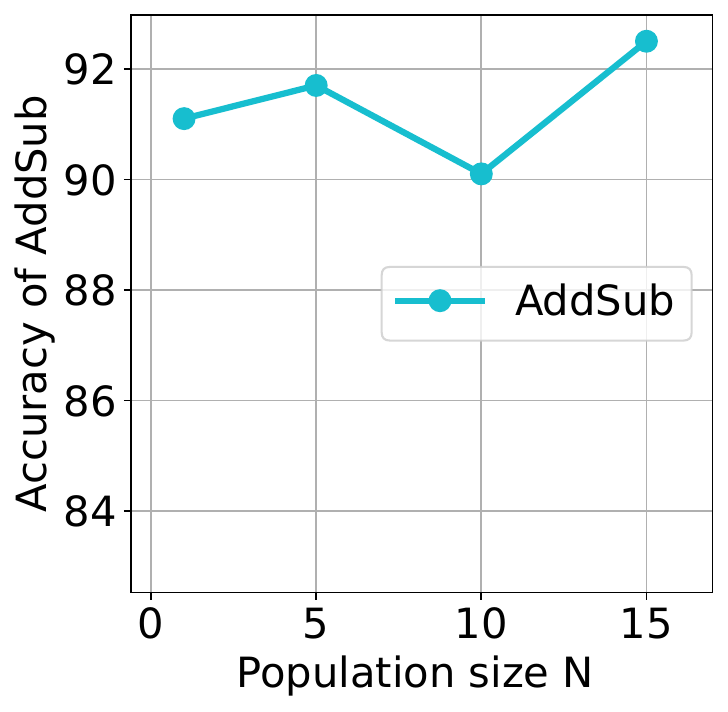}
\label{addsub}
\end{minipage}%
}
\subfigure{
\begin{minipage}[t]{0.45\linewidth}
\centering
\includegraphics[scale=0.27]{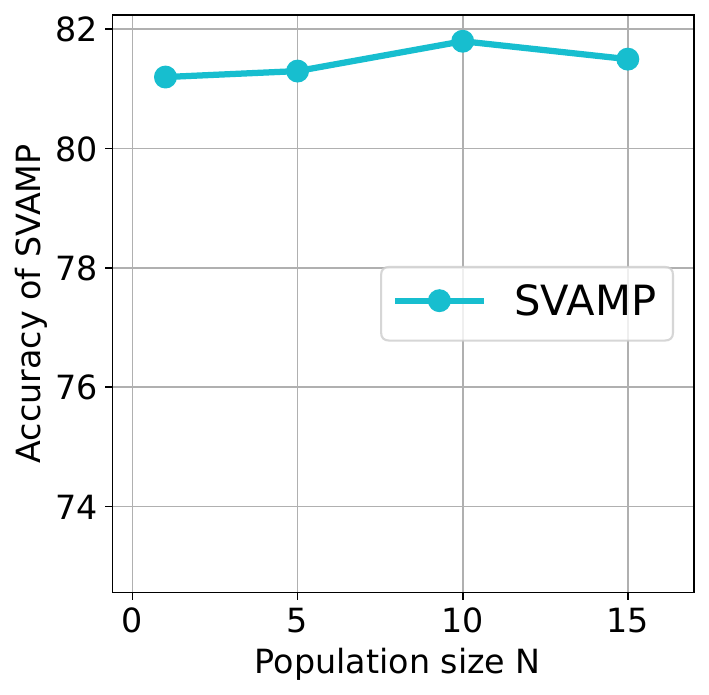}
\label{svamp}
\end{minipage}%
}
\subfigure{
\begin{minipage}[t]{0.45\linewidth}
\centering
\includegraphics[scale=0.27]{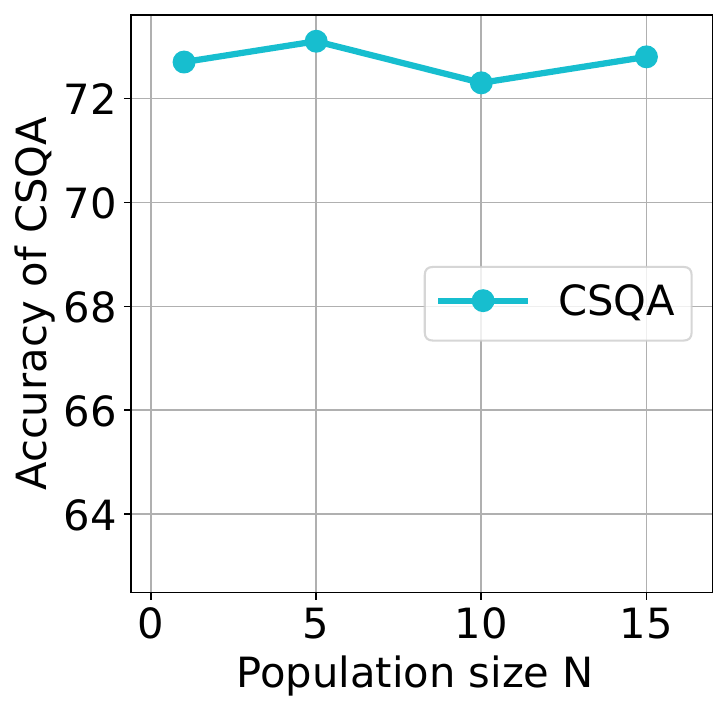}
\label{csqa}
\end{minipage}%
}
\caption{Results of different population size $N$ measured on four math reasoning datasets with GPT-3.5-turbo.}
\label{figure_popsize}
\end{figure}

In our prior experiments, we strategically employ the EoT prompting method to facilitate a singular round of crossover and mutation operations, aiming to optimize inference speed. In this context, our objective is to systematically verify the relationship between the number of our EoT promptings (i.e., represented as the population size $N$) generated during multiple rounds of crossover and mutation operations and the ensuing model performance. As depicted in Figure \ref{figure_popsize}, we conduct the experiments across four arithmetic reasoning datasets, including SingleEq, AddSub, SVAMP, and CSQA. The results manifest a discernible positive correlation, wherein an increased quantity of CoT promptings (i.e., a larger population size $N$) corresponds to a consistent enhancement in the model's performance. Thus, in scenarios where inference speed is either of lesser concern or can be overlooked, our EoT prompting affords substantial performance gains. This empirical evidence substantiates the efficacy of our proposed approach.

\subsection{Effect of Initialization Prompts}
To assess the impact of varied initializations of CoT prompting on the ensuing quality of generated EoT prompting, we conduct a series of experiments to systematically investigate the influence of EoT prompting instructions. As illustrated in Table \ref{table_trigger}, the experiments encompass four arithmetic reasoning datasets: AddSub, SVAMP, AQuA, and GSM8K. P1 designates the prompt employed by the zero-shot CoT prompting method, while P2, P3, and P4 signify the prompts integral to our proposed method. Notably, it is observed that the EoT prompting instruction utilized in P4 exhibits superior performance, surpassing the previously employed P3 in antecedent experiments. This observation underscores the potential for leveraging evolutionary algorithms to generate CoT promptings for each instance, thereby warranting further exploration. We show the example outputs of different reasoning tasks in Appendix \ref{example_output}.
\begin{table*}[!htbp]
\centering
\caption{Performance comparison of trigger sentences measured on four math reasoning datasets with GPT-3.5-turbo.}
\resizebox{\textwidth}{!}{%
\renewcommand{\arraystretch}{1.3}
\begin{tabular}{m{1cm}<{\centering}m{0.9\textwidth}| m{1cm}<{\centering}m{1cm}<{\centering}m{1cm}<{\centering}m{1cm}<{\centering}}
\toprule
No. &Trigger Sentence & AddSub & SVAMP & AQuA &GSM8K
  \\ \midrule
P1 & Let's think step by step.    & 86.5 & 79.0 & 55.3  &75.3 \\ \midrule
P2 & Below are the two Prompts:
Prompt 1: Let's think step by step.
Prompt 2: Let's first understand the problem and carefully extract all relevant variables and their corresponding numerals. Then, Let's calculate intermediate variables (pay attention to correct numerical calculation and commonsense) and solve the problem step by step carefully. I'd like you to follow the instruction step-by-step to generate a new prompt:
1. \textcolor{orange}{Crossover the two prompts and generate a new prompt.} 
2. \textcolor{orange}{Mutate the prompt generated in Step 1 and generate a new prompt.}
3. \textcolor{orange}{Select the best-generated prompt directly to think, rewrite}, solve the problem step by step, and show the answer.  & 90.6    & 80.4 & 59.7  & 75.9  \\ \midrule
P3 & \textcolor{orange}{I'd like you to first crossover and mutate the following two prompts to generate a new prompt:} Prompt 1: Let's first understand the problem and carefully extract all relevant variables and their corresponding numerals. Then, Let's calculate intermediate variables (pay attention to correct numerical calculation and commonsense) and solve the problem step by step carefully. Prompt 2: Let's first understand the problem and solve the problem step by step. Finally, \textcolor{orange}{select the best-generated prompt to think, rewrite}, solve the problem step by step, and show the answer. & 91.1   & 81.2 & 62.2  &76.8 \\ \midrule
P4 & I'd like you to follow the instructions step-by-step to solve the problem step by step, and show the answer. 1. \textcolor{cyan}{Crossover the following prompts and generate a new prompt:} Prompt 1: Let's first understand the problem and carefully extract all relevant variables and their corresponding numerals. Then, Let's calculate intermediate variables (pay attention to correct numerical calculation and common sense) and solve the problem step by step carefully. Prompt 2: Let's first understand the problem and solve the problem step by step. 2. \textcolor{cyan}{ Mutate the crossover prompt in Step 1 to generate the final prompt.} 3. \textcolor{cyan}{Apply the final prompt in Step 2 to think, rewrite}, solve the problem step by step, and show the answer. &\textbf{91.6}    & \textbf{82.8}   & \textbf{63.1} &\textbf{77.1} \\ \bottomrule
\end{tabular}
}
\label{table_trigger}
\end{table*}

\section{Related Work}
\noindent\textbf{LLMs and Prompting} 
With the increasing model complexity and the scale of parameters, LLMs have unlocked emerging capabilities, notably in-context learning (ICL) \cite{Brown2020LanguageMA}. The ICL strategy directly incorporates demonstrations into manually crafted prompts, enabling LLMs to perform exceptionally well without requiring task-specific fine-tuning. Recently, researchers have proposed continuous prompt tuning \cite{li-liang-2021-prefix,lester2021power,Liu2021GPTUT} to overcome challenges in discrete prompt searching. For instance, \citet{DBLP:journals/corr/abs-2204-04497} and \citet{DBLP:journals/talip/JinL0Z23} seek suitable prompts for each instance by learning continuous prompt information relevant to the instance. However, these methods require fine-tuning the parameters of the entire model, which is not friendly for LLMs. In contrast, our EoT prompting seeks suitable prompt information for each instance in a discrete space, avoiding fine-tuning the parameters of the entire model while maintaining good interpretability and robustness.

\noindent\textbf{LLMs and Optimization Algorithms }
Recent research has seen a flourishing exploration of treating LLMs as optimizers \cite{anonymous2024large,Liu2023LargeLM,Meyerson2023LanguageMC}. Relying on their formidable capabilities, some recent endeavors have demonstrated impressive performance in tasks such as neural network search \cite{Chen2023EvoPromptingLM}, mathematical problem-solving \cite{RomeraParedes2023MathematicalDF}, and various other domains by integrating LLMs with evolutionary algorithms \cite{Guo2023ConnectingLL,Mouret2024LargeLM,anonymous2024,hollmann2023large}. In our work, we pioneer the application of considering LLMs as part of evolutionary algorithms, specifically applying our EoT prompting to CoT reasoning, yielding favorable results across diverse tasks. 

\noindent\textbf{Chain-of-Thought Prompting}
Built upon in-context learning \cite{Brown2020LanguageMA}, the recently introduced CoT prompting \cite{NEURIPS2022_8bb0d291,NEURIPS2022_9d560961,wang2023selfconsistency} significantly enhances the reasoning capabilities of LLMs. CoT prompting not only deepens the model's understanding of subtle questions and their underlying logic but also generates a series of explicit reasoning steps. Subsequent works \cite{wang-etal-2023-plan,schaeffer2023are,zhang2023automatic,Xu2023ReReadingIR} have proposed different approaches to address complex problems. Our EoT prompting, by treating LLMs as evolutionary optimizers and generating distinct discrete CoT promptings for each instance, demonstrates superior performance across various reasoning problems.
\section{Conclusion}
In this paper, we have introduced a novel zero-shot CoT prompting called zero-shot EoT prompting. EoT prompting generates diverse CoT promptings tailored to specific instances within a task through evolutionary algorithms. The proposed method surpasses existing zero-shot CoT, PS+ prompting, and RE2 prompting methods across various reasoning datasets, demonstrating notable performance, especially in arithmetic and symbolic reasoning. Extensive experiments and analyses validate the effectiveness of zero-shot EoT prompting, showcasing its potential to enhance LLMs' reasoning capabilities. We believe there is considerable potential for refining the application of evolutionary algorithms based on LLMs to enhance model reasoning capabilities. We leave this for further exploration in the future.
\section*{Limitations}
In our proposed method, we have integrated core elements of evolutionary algorithms to leverage the capabilities of large language models for chain-of-thought reasoning. Notably, certain evolutionary algorithms, such as differential evolution, still need to be explored in our current experimentation and could be deferred for investigation in future endeavors. Our preliminary experiments are exclusively conducted using GPT-3.5-turbo and GPT-4. Considering the substantial costs associated with API usage, we intend to broaden the validation of our proposed method across a more extensive range of large language models in subsequent stages, aiming to enhance the generalizability and robustness of our method, ensuring its applicability across various language models and further validating its efficacy. 
Moreover, we do not evaluate our proposed EoT prompting under the few-shot setting because of the substantial costs associated with API usage. We leave this for further exploration in the future.

\bibliography{anthology}
\clearpage
\appendix
\onecolumn
\section{Details of Experimental Setup}
\subsection{Datasets}\label{dataset}
Table \ref{table-dataset} shows the statistics of datasets used in our experiment.
\begin{table*}[!htbp]
\centering
\caption{Details of datasets evaluated in our experiment.}
\begin{tabular}{@{}lccccc@{}}
\toprule
No. &Dataset & Samples & Avg Words & Answer Format  & Domain \\ \midrule
1 & SingleEq  & 508 & 27.4 & Number & Math  \\
2 & AddSub   & 395 & 31.5 & Number & Math \\
3 & GSM8K  & 1319 & 46.9 & Number &  Math \\
4 & MultiArith  & 600 & 31.8 & Number & Math\\
5 & SVAMP  & 1000 & 31.8 & Number & Math  \\
6 & AQuA  & 254 & 51.9 & Option & Math  \\
7 & CommonsenseQA   & 1221 & 27.8 & Option & Commonsense \\
8 & StrategyQA  & 2290 & 9.6 & Yes/No & Commonsense \\
9 & Coin Flip  & 500 & 37.0 & Yes/No & Symbolic\\
10 & Last Letters  & 1000 & 15.0 & String & Symbolic  \\
\midrule
\end{tabular}
\label{table-dataset}
\end{table*}
\subsection{Answer Extraction Prompts}\label{answer_ext}
Table \ref{table-extraction} shows a list of answer extraction prompts used in our experiments.
\begin{table*}[!htbp]
\centering
\caption{Answer extraction prompts used in our experiments across all baselines.}
\begin{tabular}{@{}lll@{}}
\toprule
No & Dataset & Answer Extraction Prompts \\ \midrule
1 & SingleEq  & Therefore, the answer (arabic numerals) is  \\
2 & AddSub   & Therefore, the answer (arabic numerals) is \\
3 & GSM8K  & Therefore, the answer (arabic numerals) is \\
4 & MultiArith  & Therefore, the answer (arabic numerals) is\\
5 & SVAMP  & Therefore, the answer (arabic numerals) is  \\
6 & AQuA  & Therefore, among A through E, the answer is \\
7 & CommonsenseQA   & Therefore, among A through E, the answer is \\
8 & StrategyQA  & Therefore, the answer (Yes or No) is \\
9 & Coin Flip  & Therefore, the answer (Yes or No) is \\
10 & Last Letters  & Therefore, the answer is   \\
\midrule
\end{tabular}
\label{table-extraction}
\end{table*}

\section{Example Outputs of Zero-shot EoT}\label{example_output}
Tables \ref{table_letters} to \ref{table_gsm8k} list example outputs generated by our proposed method Zero-shot EoT prompting for several datasets, including Last Letters, SVAMP, AQuA, AddSub, MultiArith, SingleEq, and GSM8K.
\begin{table*}[ht!]
\caption{Example output of Zero-shot-EoT for Last\_letters}
\label{table_letters}
\resizebox{\textwidth}{!}{
\begin{tabular}{@{}p{\textwidth}@{}}
\toprule
\textbf{Q:} Take the last letters of each words in \"Sterling Jenifer Patsy Denise\"  and concatenate them.\\
\textbf{A:} For the following the two Prompts: \\
Prompt 1: Let's think step by step. \\ Prompt 2: Let's first understand the problem and carefully extract all relevant variables and their corresponding numerals. Then, Let's calculate intermediate variables (pay attention to correct numerical calculation and commonsense) and solve the problem step by step carefully. \\ I'd like you to follow the instruction step-by-step to generate new prompts: \\ 1. Crossover the two prompts and generate a new prompt. \\
2. Mutate the prompt generated in Step 1 and generate a new prompt. \\ 3. Select the best-generated prompt directly to think, rewrite, solve the problem step by step, and show the answer. \\
\textbf{Output:} \\ Prompt 1: Let's carefully analyze the situation. Given the words "Sterling," "Jenifer," "Patsy," and "Denise," what is the specific task we are asked to perform with the last letters of each word? Identify the relevant variables and outline the steps to concatenate the last letters.\\ Prompt 2: Consider the problem step by step. Identify the words in the given set - "Sterling," "Jenifer," "Patsy," and "Denise." Extract the last letters of each word and outline the process to concatenate them. Pay attention to correct extraction and concatenation procedures.\\
Prompt 3: Let's break down the problem systematically. List the words "Sterling," "Jenifer," "Patsy," and "Denise." Determine the last letter of each word and devise a method to concatenate them. Ensure precision in extracting and combining the letters.\\
Prompt 4: Recap the problem understanding. Specify the words involved - "Sterling," "Jenifer," "Patsy," and "Denise." Clearly define the steps to extract the last letters from each word and concatenate them. Ensure accuracy in the extraction process and the final concatenation.\\

Now, let's proceed with solving the problem step by step using the information provided in the prompts:\\

Words: Sterling, Jenifer, Patsy, Denise\\

Last letters: g (Sterling), r (Jenifer), y (Patsy), e (Denise)\\

Concatenate the last letters: grye\\

So, the result of concatenating the last letters of each word is "grye." \\ \bottomrule
\end{tabular}
}
\end{table*}

\begin{table*}[ht!]
\caption{Example output of Zero-shot-EoT for SVAMP}
\label{table_svamp}
\resizebox{\textwidth}{!}{
\begin{tabular}{@{}p{\textwidth}@{}}
\toprule
\textbf{Q:}  Jack received 10 emails in the morning, 5 emails in the afternoon and 4 emails in the evening. How many more emails did Jack receive in the afternoon than in the evening? \\
\textbf{A:} I'd like you to crossover and mutate the following two prompts to generate a new prompt: \\ Prompt 1: Let's first understand the problem and carefully extract all relevant variables and their corresponding numerals. Then, Let's calculate intermediate variables (pay attention to correct numerical calculation and commonsense) and solve the problem step by step carefully. \\ Prompt 2: Let's first understand the problem and devise a plan to solve the problem step by step. \\ Finally, select the best-generated prompt to think, rewrite, solve the problem step by step, and show the answer.\\
\textbf{Output:} Best-Generated Prompt:
"Analyze the scenario where Jack received 10 emails in the morning, 5 emails in the afternoon, and 4 emails in the evening. Let's first understand the problem and carefully extract relevant variables and their corresponding numerals. Devise a step-by-step plan to solve the problem, paying attention to correct numerical calculation and commonsense. Determine the number of additional emails Jack received in the afternoon compared to the evening." \\
Now, let's solve the problem step by step: \\
Jack received 5 emails in the afternoon and 4 emails in the evening. \\ Calculate the difference: 5-4=1. \\
Therefore, Jack received 1 more email in the afternoon than in the evening.
\\ \bottomrule
\end{tabular}
}
\end{table*}

\begin{table*}[ht!]
\caption{Example output of Zero-shot-EoT for AQuA}
\label{table_aqua}
\resizebox{\textwidth}{!}{
\begin{tabular}{@{}p{\textwidth}@{}}
\toprule
\textbf{Q:} A rope 20 meters long is cut into two pieces. If the length of one piece of rope is 3 meters shorter than the length of the other, what is the length, in meters, of the longer piece of rope? Answer Choices: (A) 7.5 (B) 8.9 (C) 9.9 (D) 11.5 (E) 11.7. \\
\textbf{A:} I'd like you to follow the instructions step-by-step to solve the problem step by step and show the answer.\\ 
1. Crossover the following prompts and generate a new prompt: \\ Prompt 1: Let's first understand the problem and carefully extract all relevant variables and their corresponding numerals. Then, Let's calculate intermediate variables (pay attention to correct numerical calculation and common sense) and solve the problem step by step carefully.\\ Prompt 2: Let's first understand the problem and solve the problem step by step. \\
2. Mutate the crossover prompt in Step 1 to generate the final prompt.\\
3. Apply the final prompt in Step 2 to think, rewrite, solve the problem step by step, and show the answer.\\
\textbf{Output:} Crossover Prompt:\\
Prompt 1: Let's first understand the problem and carefully extract all relevant variables and their corresponding numerals. Then, Let's calculate intermediate variables (pay attention to correct numerical calculation and common sense) and solve the problem step by step carefully.\\
Prompt 2: Let's first understand the problem and solve the problem step by step.\\
Mutated Crossover Prompt:\\
Let's carefully extract all relevant variables and their corresponding numerals from the problem. Then, calculate intermediate variables with attention to correct numerical calculation and common sense. Finally, solve the problem step by step.\\
Applying the Final Prompt:\\
Given:\\
Total length of the rope = 20 meters\\
Length of one piece of rope = x meters\\
Length of the other piece of rope = (x + 3) meters\\
We need to find the length of the longer piece of rope.
Breaking down the information:\\
x + (x + 3) = 20 (since the total length is the sum of the two pieces)\\
Combine like terms: 2x + 3 = 20\\
Subtract 3 from both sides: 2x = 17\\
Divide by 2: x = 8.5\\
The length of the longer piece (x + 3) is 8.5 + 3 = 11.5 meters.\\
Therefore, the correct answer is (D) 11.5 meters.
\\ \bottomrule
\end{tabular}
}
\end{table*}

\begin{table*}[ht!]
\caption{Example output of Zero-shot-EoT for AddSub}
\label{table_addsub}
\resizebox{\textwidth}{!}{
\begin{tabular}{@{}p{\textwidth}@{}}
\toprule
\textbf{Q:} There are 54 scissors in the drawer . Keith placed 22 scissors in the drawer . How many scissors are now there in all? \\
\textbf{A:} I'd like you to follow the instructions step-by-step to solve the problem step by step and show the answer.\\ 
1. Crossover the following prompts and generate a new prompt: \\ Prompt 1: Let's first understand the problem and carefully extract all relevant variables and their corresponding numerals. Then, Let's calculate intermediate variables (pay attention to correct numerical calculation and common sense) and solve the problem step by step carefully.\\ Prompt 2: Let's first understand the problem and solve the problem step by step. \\
2. Mutate the crossover prompt in Step 1 to generate the final prompt.\\
3. Apply the final prompt in Step 2 to think, rewrite, solve the problem step by step, and show the answer.\\
\textbf{Output:} Crossover Prompt:\\
Let's first understand the problem and carefully extract all relevant variables and their corresponding numerals. Then, let's calculate intermediate variables (pay attention to correct numerical calculation and common sense) and solve the problem step by step carefully.\\
Mutated Prompt:\\
Let's carefully extract all relevant variables and their corresponding numerals from the problem. Calculate intermediate variables with attention to correct numerical calculations and common sense. Solve the problem step by step.
\\Final Prompt:\\
Let's carefully extract all relevant variables and their corresponding numerals from the problem. Calculate intermediate variables with attention to correct numerical calculations and common sense. Solve the problem step by step.
\\ Now, let's apply the final prompt to solve the problem:\\
Given:\\
Total number of scissors in the drawer = 54 \\
Scissors placed by Keith = 22 \\
We need to find the total number of scissors after Keith placed 22 scissors in the drawer.\\
Step 1: Understand the problem and extract variables
Total scissors in the drawer (initial) = 54
Scissors placed by Keith = 22 \\
Step 2: Calculate intermediate variables
Total scissors after Keith placed = Total scissors in the drawer (initial) + Scissors placed by Keith
Total scissors after Keith placed = 54 + 22 \\
Step 3: Solve the problem step by step
Total scissors after Keith placed = 76\\
Answer:
After Keith placed 22 scissors in the drawer, there are now 76 scissors in total.
\\ \bottomrule
\end{tabular}
}
\end{table*}

\begin{table*}[ht!]
\caption{Example output of Zero-shot-EoT for MultiArith}
\label{table_multiarith}
\resizebox{\textwidth}{!}{
\begin{tabular}{@{}p{\textwidth}@{}}
\toprule
\textbf{Q:} A store had 27 coloring books in stock. They ended up putting them on sale and getting rid of 6 of them. The put the ones they still had onto shelves with 7 on each shelf. How many shelves did they use? \\
\textbf{A:} I'd like you to follow the instructions step-by-step to solve the problem step by step and show the answer.\\ 
1. Crossover the following prompts and generate a new prompt: \\ Prompt 1: Let's first understand the problem and carefully extract all relevant variables and their corresponding numerals. Then, Let's calculate intermediate variables (pay attention to correct numerical calculation and common sense) and solve the problem step by step carefully.\\ Prompt 2: Let's first understand the problem and solve the problem step by step. \\
2. Mutate the crossover prompt in Step 1 to generate the final prompt.\\
3. Apply the final prompt in Step 2 to think, rewrite, solve the problem step by step, and show the answer.\\
\textbf{Output:} Crossover Prompt: \\
Prompt 1: Let's first understand the problem and carefully extract all relevant variables and their corresponding numerals. Then, let's calculate intermediate variables (pay attention to correct numerical calculation and common sense) and solve the problem step by step carefully.\\
Prompt 2: Let's first understand the problem and solve the problem step by step.\\
Mutated Crossover Prompt:\\
Extract relevant variables and their corresponding numerals from the problem. Then, calculate intermediate variables with attention to correct numerical calculation and common sense. Finally, solve the problem step by step.\\
Applying the Final Prompt:\\
Given:\\
Initial stock of coloring books = 27\\
Coloring books sold = 6\\
Coloring books remaining = 27 - 6 = 21\\
Number of coloring books on each shelf = 7\\
We need to find the number of shelves used to display the remaining coloring books.\\
Calculation:\\
Shelves used = Remaining coloring books / Coloring books per shelf\\
Shelves used = 21 / 7\\
Simplifying:
Shelves used = 3\\
Therefore, the store used 3 shelves to display the remaining coloring books.
\\ \bottomrule
\end{tabular}
}
\end{table*}

\begin{table*}[ht!]
\caption{Example output of Zero-shot-EoT for SingleEq}
\label{table_singleeq}
\resizebox{\textwidth}{!}{
\begin{tabular}{@{}p{\textwidth}@{}}
\toprule
\textbf{Q:} Dan has 29 violet balloons. Tim has 7 times more violet balloons than Dan. How many violet balloons does Tim have? \\
\textbf{A:} I'd like you to follow the instructions step-by-step to solve the problem step by step and show the answer.\\ 
1. Crossover the following prompts and generate a new prompt: \\ Prompt 1: Let's first understand the problem and carefully extract all relevant variables and their corresponding numerals. Then, Let's calculate intermediate variables (pay attention to correct numerical calculation and common sense) and solve the problem step by step carefully.\\ Prompt 2: Let's first understand the problem and solve the problem step by step. \\
2. Mutate the crossover prompt in Step 1 to generate the final prompt.\\
3. Apply the final prompt in Step 2 to think, rewrite, solve the problem step by step, and show the answer.\\
\textbf{Output:} Crossover Prompt:\\
Prompt 1: Let's first understand the problem and carefully extract all relevant variables and their corresponding numerals. Then, let's calculate intermediate variables (pay attention to correct numerical calculation and common sense) and solve the problem step by step carefully.\\
Prompt 2: Let's first understand the problem and solve the problem step by step.\\
Mutated Crossover Prompt:\\
Extract relevant variables and their corresponding numerals from the problem. Then, calculate intermediate variables with attention to correct numerical calculation and common sense. Finally, solve the problem step by step.\\
Applying the Final Prompt:\\
Given:\\
Number of violet balloons Dan has = 29\\
Tim has 7 times more violet balloons than Dan\\
We need to find the number of violet balloons Tim has.\\
Calculation:\\
Number of violet balloons Tim has = 7 * Number of violet balloons Dan has\\
Number of violet balloons Tim has = 7 * 29\\
Simplifying:
Number of violet balloons Tim has = 203\\
Therefore, Tim has 203 violet balloons.
\\ \bottomrule
\end{tabular}
}
\end{table*}

\begin{table*}[ht!]
\caption{Example output of Zero-shot-EoT for GSM8K}
\label{table_gsm8k}
\resizebox{\textwidth}{!}{
\begin{tabular}{@{}p{\textwidth}@{}}
\toprule
\textbf{Q:} A merchant wants to make a choice of purchase between 2 purchase plans: jewelry worth \$5,000 or electronic gadgets worth \$8,000. His financial advisor speculates that the jewelry market will go up 2.5\% while the electronic gadgets market will rise 1.2\% within the same month. If the merchant is looking to maximize profit at the end of this month by making a choice, how much profit would this be? \\
\textbf{A:} I'd like you to follow the instructions step-by-step to solve the problem step by step and show the answer.\\ 
1. Crossover the following prompts and generate a new prompt: \\ Prompt 1: Let's first understand the problem and carefully extract all relevant variables and their corresponding numerals. Then, Let's calculate intermediate variables (pay attention to correct numerical calculation and common sense) and solve the problem step by step carefully.\\ Prompt 2: Let's first understand the problem and solve the problem step by step. \\
2. Mutate the crossover prompt in Step 1 to generate the final prompt.\\
3. Apply the final prompt in Step 2 to think, rewrite, solve the problem step by step, and show the answer.\\
\textbf{Output:} Crossover Prompt:\\
Prompt 1: Let's first understand the problem and carefully extract all relevant variables and their corresponding numerals. Then, let's calculate intermediate variables (pay attention to correct numerical calculation and common sense) and solve the problem step by step carefully.\\
Prompt 2: Let's first understand the problem and solve the problem step by step.\\
Mutated Crossover Prompt:\\
Extract relevant variables and their corresponding numerals from the problem. Then, calculate intermediate variables with attention to correct numerical calculation and common sense. Finally, solve the problem step by step.\\
Applying the Final Prompt:\\
Given:\\
Initial investment in jewelry = \$5,000\\
Initial investment in electronic gadgets = \$8,000\\
Speculated increase in the jewelry market = 2.5\%\\
Speculated increase in the electronic gadgets market = 1.2\%\\
We need to find the profit for each investment option.\\
Calculation:\\
Profit from jewelry investment = Initial investment * (Speculated increase / 100)\\
Profit from jewelry investment = \$5,000 * (2.5 / 100)\\
Simplifying:
Profit from jewelry investment = \$125\\
Profit from electronic gadgets investment = Initial investment * (Speculated increase / 100)\\
Profit from electronic gadgets investment = \$8,000 * (1.2 / 100)\\
Simplifying:
Profit from electronic gadgets investment = \$96\\
Now, compare the profits:\\
Profit from jewelry investment = \$125\\
Profit from electronic gadgets investment = \$96\\
The merchant should choose the jewelry investment to maximize profit, and the profit would be \$125.
\\ \bottomrule
\end{tabular}
}
\end{table*}
\end{document}